\documentclass[lettersize,journal]{IEEEtran}
\usepackage{amsmath,amsfonts}
\usepackage{algorithmic}
\usepackage{algorithm}
\usepackage{array}
\usepackage[caption=false,font=normalsize,labelfont=sf,textfont=sf]{subfig}
\usepackage{textcomp}
\usepackage{stfloats}
\usepackage{url}
\usepackage{verbatim}
\usepackage{graphicx}
\usepackage{multirow}
\usepackage{cite}
\begin{document}

\title{Towards Lifelong Few-Shot Customization of Text-to-Image Diffusion}

\author{Nan Song, Xiaofeng Yang, Ze Yang, Guosheng Lin
\thanks{N. Song, X. Yang, Z. Yang, and G. Lin are with the Nanyang Technological University, Singapore. E-mail: \{nan001, xiaofeng001,
ze001\}@e.ntu.edu.sg, gslin@ntu.edu.sg. Corresponding to: Guosheng Lin.}
}

\markboth{Journal of \LaTeX\ Class Files,~Vol.~14, No.~8, August~2021}%
{Shell \MakeLowercase{\textit{et al.}}: A Sample Article Using IEEEtran.cls for IEEE Journals}

\IEEEpubid{0000--0000/00\$00.00~\copyright~2021 IEEE}

\maketitle

\begin{abstract}
Lifelong few-shot customization for text-to-image diffusion aims to continually generalize existing models for new tasks with minimal data while preserving old knowledge.
Current customization diffusion models excel in few-shot tasks but struggle with catastrophic forgetting problems in lifelong generations. In this study, we identify and categorize the catastrophic forgetting problems into two folds:  relevant concepts forgetting and previous concepts forgetting. 
To address these challenges, we first devise a data-free knowledge distillation strategy to tackle relevant concepts forgetting. Unlike existing methods that rely on additional real data or offline replay of original concept data, our approach enables on-the-fly knowledge distillation to retain the previous concepts while learning new ones, without accessing any previous data.
Second, we develop an In-Context Generation (ICGen) paradigm that allows the diffusion model to be conditioned upon the input vision context, which facilitates few-shot generation and mitigates the issue of previous concepts forgetting.
Extensive experiments show that the proposed Lifelong Few-Shot Diffusion (LFS-Diffusion) method can produce high-quality and accurate images while maintaining previously learned knowledge.
\end{abstract}


\section{Introduction}
\IEEEPARstart{H}{uman} learning occurs in a lifelong manner, which allows us to progressively acquire and accumulate knowledge. This process enhances our proficiency in learning and problem-solving when we encounter new challenges. However, machine learning systems often encounter a phenomenon known as catastrophic forgetting~\cite{mccloskey1989catastrophic,kirkpatrick2017overcoming}. This problem occurs when these systems are given new tasks, causing difficulty in retaining knowledge from earlier tasks. Furthermore, modern deep-learning models~\cite{imagenet,resnet} require large volumes of training data, which can be both expensive and time-consuming in real-world applications. In numerous domains, data collection is inherently limited, whether in the medical realm that involves patient privacy issues or the artistic field where accessing specific artists' works is restricted. The difference between human learning and machine learning algorithms has spurred increasing attention toward incremental few-shot learning~\cite{mazumder2021few,qin2022lfpt, Zhang_2021_CVPR}. This research area seeks to create models that are capable of consistently accommodating new tasks using minimal training samples.

Lifelong few-shot generation, initially presented in LFS-GAN~\cite{seo2023lfs}, aims to develop a generative model that can create varied and authentic images with minimal training data. This approach continuously adjusts to new tasks while maintaining the ability to generate images based on prior learning. 
Recently introduced diffusion models~\cite{ho2020denoising, song2020denoising, Zhang_2023_ICCV, gu2022vector} have achieved remarkable progress in computer vision applications, significantly enhancing the performance of general generative models. In particular, the Stable Diffusion (SD) technique~\cite{rombach2022high} has enabled diffusion models to act as versatile and powerful generators, excelling in tasks such as text-to-image synthesis, image inpainting, video generation, and more.

\begin{figure*}[t]
    \centering
    \includegraphics[width=1.0\linewidth]{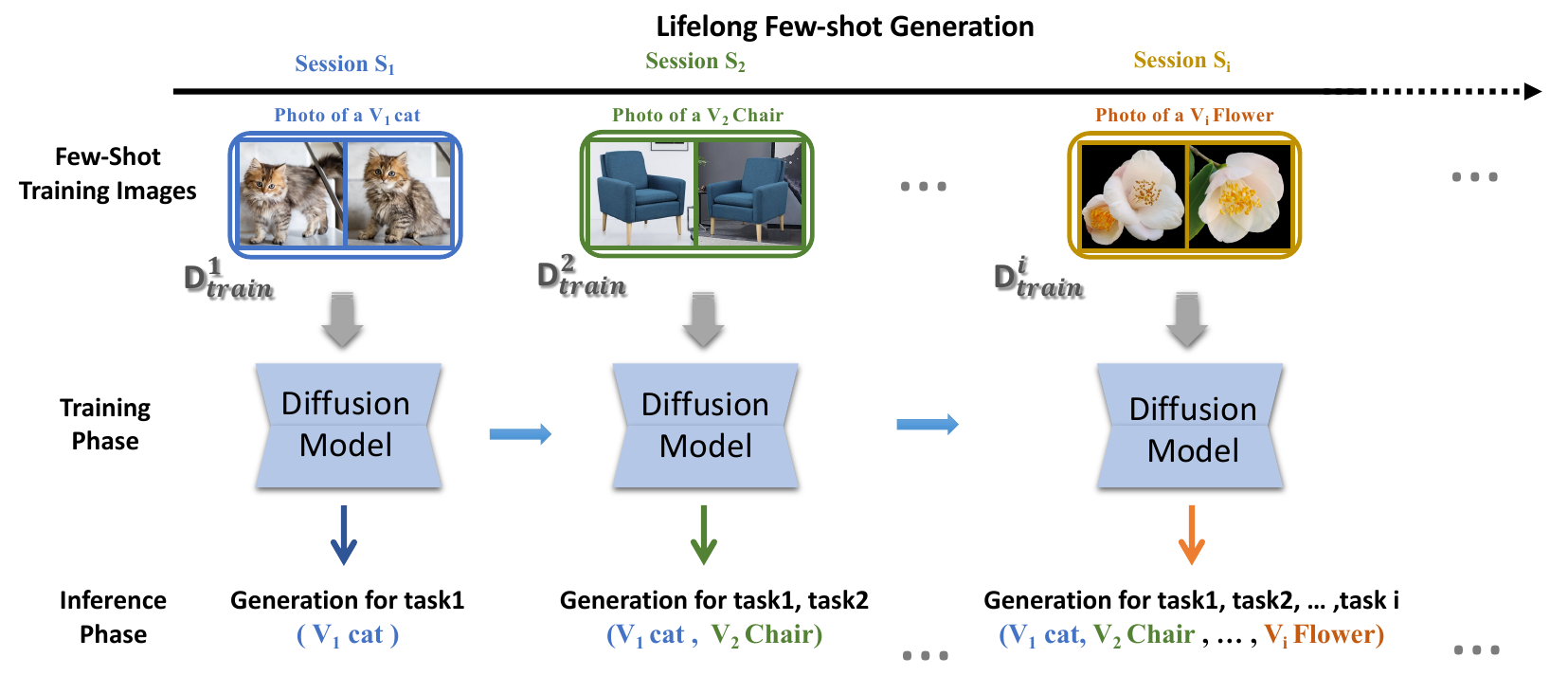}
    \caption{Lifelong few-shot text-to-image diffusion. Lifelong few-shot learning aims to continually learn multiple sessions of tasks without forgetting the previous ones. In each session, the training dataset contains a few images of a new concept. The model learns the new concept during the training phase, enabling it to generate both the novel concept from the current session and previously learned concepts from past sessions. Current customization diffusion models suffer from catastrophic forgetting problems in lifelong generation tasks. }
    \label{fig:teaser}
\end{figure*}

\begin{figure}[t]
	\centering
	\includegraphics[width=0.88\linewidth]{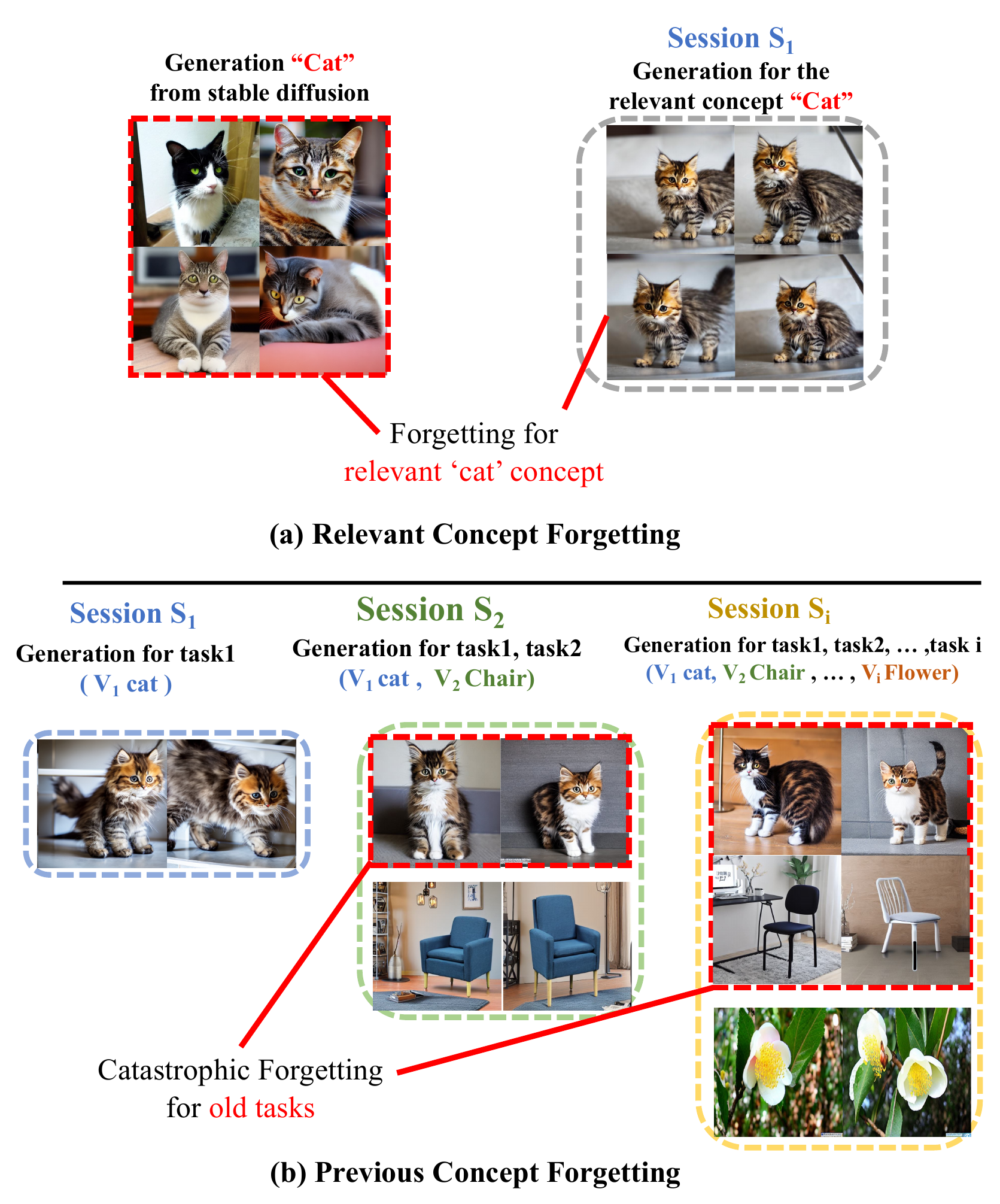}
	\caption{We identify two catastrophic forgetting problems in lifelong generations: Relevant Concepts Forgetting (RCF) and Previous Concepts forgetting (PCF). Left: RCF refers to forgetting the relevant concepts related to the new concepts. After training on session 1 ``$V_1$ cat", the model tends to generate relevant concept ``Cat" all similar to ``$V_1 $ cat". Right: PCF denotes forgetting the concepts learned in the previous sessions. For instance, in session $i$, the previous concept ``blue $V_2$ chair" fails to be generated when learning a new concept ``$V_i$ flower". }
	\label{fig:rcf_pcf}
\end{figure}

In order to leverage the potential of SD for learning personalized new concepts, recent works~\cite{ruiz2023dreambooth,kumari2022customdiffusion, Tewel2023KeyLockedRO} have concentrated on customizing text-to-image diffusion models with few-shot personalized images through fine-tuning.
Current customization diffusion models perform well in few-shot tasks but suffer from catastrophic forgetting problems under lifelong scenarios.
\textbf{We identify two catastrophic forgetting problems in lifelong
generations: Relevant Concepts Forgetting (RCF) and Previous Concepts Forgetting (PCF).} As shown in Figure~\ref{fig:rcf_pcf}. RCF refers to forgetting the relevant concepts related to the new concepts, whereas PCF denotes forgetting the concepts learned in the previous sessions. 
To illustrate, consider sequential customization scenarios involving two distinct entities: a new type of cat, termed ``$V_1$ cat'', and a novel table design, labeled ``$V_2$ table''. RCF is observed when the pre-trained Text-to-Image (T2I) models, upon being trained with ``$V_1$ cat'' data, exhibit diminished recall of other ``cat'' concepts previously learned, as shown in Fig~\ref{fig:rcf_pcf}(a). Conversely, PCF manifests as the loss of the ``$V_1$ cat'' concept from the model’s memory following its training on the ``$V_2$ table'' data. These issues arise when models are exposed to new tasks and fail to retain the knowledge of previously learned concepts or relevant features, which is critical for real-world applications where continuous learning with minimal data is required

\IEEEpubidadjcol  To explore alternative solutions, we evaluated various methodologies. Firstly, the traditional knowledge distillation techniques~\cite{Zhai_2019_ICCV, gao2023ddgr}, which rely heavily on data storage or replay mechanisms, were found to be memory and computationally expensive. This raised concerns about scalability, especially in environments with data privacy or availability constraints, such as healthcare or copyright-sensitive domains. Secondly, existing lifelong learning approaches in generative models, such as Elastic Weight Consolidation~\cite{kirkpatrick2017overcoming} and memory-replay-based methods~\cite{wu2018memory}, showed promise but faced challenges in low-data scenarios, leading to catastrophic forgetting and poor performance when adapting to new tasks incrementally. Thirdly, the in-context learning technics demonstrated great potential in NLP tasks by leveraging context during inference to improve task adaptability. However, its application in vision-based tasks, particularly in diffusion models, remains underexplored.

Our research is motivated by these gaps and seeks to propose a solution that combines data-free knowledge distillation and in-context generation to address RCF and PCF efficiently. By integrating these strategies, we aim to push the boundaries of lifelong few-shot learning in diffusion models, enabling continual learning without the need for vast data storage or additional memory overhead.

To solve RCF, 
we introduce an innovative data-free knowledge distillation technique to address the RCF issue without necessitating additional offline-generated images or human-collected images. Specifically, we augment the customization model through the distillation of a running diffusion inference process using relevant concepts. This distillation method effectively eliminates the need for offline real image data and is notably more efficient compared to prior approaches.

To overcome previous concepts forgetting (PCF), we resort to in-context learning, which is widely used in neural language processing(NLP) problems.
It enables large language models to quickly adapt to diverse tasks with limited prompts and examples~\cite{gpt3,xie2021explanation}.  Motivated by this, we introduce In-Context Generation (ICGen) to enhance the model’s performance by incorporating vision context during inference. ICGen use vision in-context learning to improve few-shot learning performance and address the issue of forgetting previously acquired knowledge. We create a diffusion model that is conditioned on both textual and visual context. By incorporating visual context with random noise, we guide the denoising process without requiring additional training.

The potential real-world applications of LFS-Diffusion include personalized image generation, continual learning for creative tasks, and scenarios where minimal training data is available, such as medical imaging or data-scarce artistic customizations.

We conduct extensive experiments on benchmark datasets to validate the effectiveness of our proposed approach. The key contributions of this study are outlined as follows:
\begin{itemize}
    \item We identify two catastrophic forgetting problems in lifelong customization of diffusion models: Relevant Concepts Forgetting (RCF) and Previous Concepts Forgetting (PCF).
    
	\item We propose data-free knowledge distillation to resolve relevant concepts forgetting. Compared to previous methods that use offline-generated or human-collected data, our method is both memory and computationally efficient.
 
	\item We introduce an In-Context Generation (ICGen) paradigm that enables the diffusion model to utilize the input vision context for in-context learning during inference. This approach enhances few-shot generation and addresses the problem of forgetting previously learned concepts.
 
	\item Experimental results reveal that our method outperforms baseline approaches and sets a new benchmark on the CustomConcept101 and DreamBooth datasets, showcasing its substantial advantages.
\end{itemize}

\section{Related Work}
\label{sec:related}

\subsection{Lifelong Image Generation}
 Several methods have been developed to address catastrophic forgetting in generative models that are trained sequentially on different tasks~\cite{seff2017continual, Zhai_2019_ICCV, Zhai_2021_CVPR, cong2020gan}. Seff et al.\cite{seff2017continual} suggested incorporating Elastic Weight Consolidation (EWC)\cite{kirkpatrick2017overcoming} into the generation loss to manage network parameters during the learning of new tasks. Memory-replay-based approaches~\cite{wu2018memory} relied on replayed samples to memorize the old tasks. Lifelong GAN~\cite{Zhai_2019_ICCV} inspired by Learning Without Forgetting (LwF)~\cite{li2017learning} employed a knowledge distillation technique that relies on additional data to prevent forgetting. HyperLifelong GAN~\cite{Zhai_2021_CVPR} introduced a shared weight matrix across different tasks and updated the shared weight when new tasks came.
While these methods have demonstrated effectiveness, they exhibit notable limitations in low-data scenarios, including severe mode collapse. In contrast, our method addresses the challenges presented by limited training data in few-shot learning by strategically utilizing vision in-context learning to improve few-shot performance.

\begin{figure*}[t]
	\centering
	\includegraphics[width=0.95\linewidth]{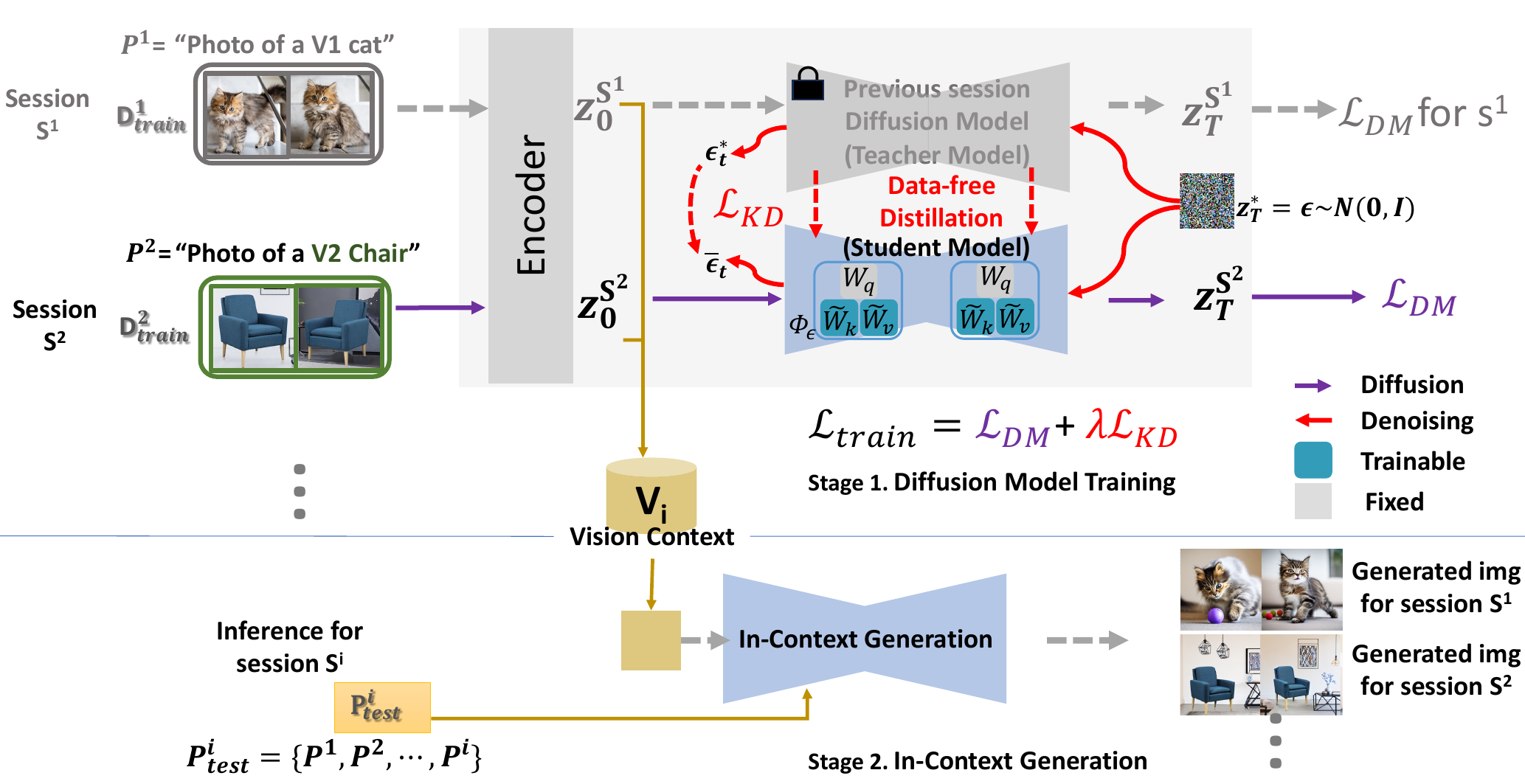}
	\caption{Our framework for lifelong few-shot text-to-image diffusion mainly includes two stages: (1) the Diffusion Model Training stage to learn the diffusion model ${\Phi}_{\epsilon}$ using the training data in the current session $\mathcal{D}_{train}^i$, (2) the In-Context Generation stage to generate images using the test prompt $P_{test}^i$ for session $S^i$ and all the previous sessions. The Diffusion Model Training
stage includes the normal diffusion process with the few-shot training data (\(L_{DM}\)) and the denoising process to distill the knowledge from the previous session to the current session (\(L_{KD}\)). Vision context $Z_0^{S^i}$ is also sampled to help the In-Context Generation. During the inference for session $S^i$, a vision context-guided text-to-image diffusion is performed without additional training. In-Context Generation can improve the few-shot learning performance and prevent the forgetting of concepts in the previous sessions.}
	\label{fig:whole}
\vskip -1em
\end{figure*}

\subsection{Text-to-Image Diffusion}
Generative models have gained research attention in recent years ~\cite{rombach2022high, saharia2022photorealistic, lian2023llmgrounded, yang2024mastering, Shen_Multimedia, Zha2012, shen2016accurate}. 
Text-to-image diffusion models have demonstrated their adaptability in various applications, particularly in image generation.  GLIDE~\cite{nichol2022glide} and Imagen~\cite{saharia2022photorealistic} both utilize a classifier-free guidance approach. GLIDE~\cite{nichol2022glide} stands out for its prowess in image generation and editing tasks. Stable Diffusion~\cite{rombach2022high}, a major advancement in latent diffusion, utilizes VQ-GAN~\cite{esser2021taming} to shift the image features from the pixel to latent to reduce the feature size for easier calculation. Concurrently, StructDiff~\cite{feng2022training} and Attend-and-Excite~\cite{chefer2023attend} tackle issues identified in synthetic images produced by pre-trained Stable Diffusion by introducing structured guidance and enhancing cross-attention mechanisms. ELITE~\cite{wei2023elite} proposes embedding the visual concept into text embedding to guide diffusion fine-tuning in new scene composition.
Lian et al.~\cite{lian2023llmgrounded} and Yang et al.~\cite{yang2024mastering} both explored using large language models to guide the diffusion model generation process. They are more focusing on the text-driven prompts. 
In contrast, our method involves encoding the visual context within the noise latent for guidance.

\textbf{Personalized Text-to-Image diffusion}: Recent efforts~\cite{ruiz2023dreambooth,gal2022image,kumari2022customdiffusion,Xie_2023_ICCV,Tewel2023KeyLockedRO,ma2023subject} have focused on leveraging the capabilities of Stable Diffusion(SD)~\cite{rombach2022high} to generate user-specific images by customizing diffusion models through fine-tuning techniques on personalized few-shot datasets. For example, DreamBooth~\cite{ruiz2023dreambooth} protects existing knowledge by employing a class-specific prior preservation loss while fine-tuning all parameters of SD. Textual Inversion~\cite{gal2022image} learns the new concepts' word embeddings and keeps the network unchanged. Custom Diffusion~\cite{kumari2022customdiffusion}, on the other hand, fine-tunes part of the network parameters in the specific attention layers, allowing for quick adaptation. Despite these efforts, all these models encounter challenges in the context of lifelong learning. Smith et al.~\cite{smith2023continualdiffusion} explore continual learning in personalized T2I by preserving past model weights and initializing custom tokens randomly. However, their approach requires saving additional model parameters, presenting a drawback. L$^2$DM~\cite{sun2023create} introduces additional long-term and short-term memory banks to facilitate lifelong customized generation.

\subsection{Few-Shot Image Generation}
In recent years, substantial progress in few-shot classification tasks~\cite{ProtoNet,feat,RelationNet,Zhang_2020_CVPR,Sun_2019_CVPR,Canfsl,Zhou_TCSVT,Jiang_TCSVT,Zhang_TCSVT,Dang_TCSVT,ZhangJing_TCSVT} has fueled a burgeoning interest in few-shot image generation. Few-shot image generation aims to produce realistic images from a few number of training samples. The FUNIT model, as outlined in \cite{liu2019few}, is designed for image-to-image translation across different domains within a few-shot context. Additionally, models like F2GAN~\cite{HongF2GAN}, LoFGAN~\cite{gu2021lofgan}, and AGE~\cite{ding2022attribute} have explored few-shot image generation through fusion-based approaches in conditional settings.

Recent studies have demonstrated the efficacy of employing networks that have been pre-trained on extensive datasets, particularly in situations where available data is extremely limited. For instance, BSA~\cite{noguchi2019image} adapts a pre-trained network through the fine-tuning of batch statistics. EWC~\cite{li2020few} leverages the Fisher information matrix to protect essential weights from changes. FreezeD~\cite{mo2020freeze} employs a distinct approach by fine-tuning a pre-trained GAN while keeping the early layers of the discriminator fixed. CDC\cite{ojha2021few-shot-gan} emphasizes the preservation of pairwise distances among generated samples, while SSRGAN~\cite{sushko2023smoothness} utilizes the smooth latent space of a pre-trained GAN, adapting it to new domains with limited data. RSSA~\cite{xiao2022few} employs a self-correlation matrix to ensure structural consistency. 
FAML~\cite{Phap_2022_TMM} conditions the feature vectors from the encoder for integration with a GAN model and uses meta-learning with few samples.

These advanced methods highlight the significant potential of utilizing pre-trained networks combined with various fine-tuning techniques to deliver compelling outcomes in few-shot image generation tasks, even under constraints of limited training data.

\subsection{In-Context Learning}
GPT-3~\cite{gpt3} introduced in-context learning, framing NLP tasks as text completions, enabling models to reason and recognize new patterns on the fly.
Flamingo~\cite{alayrac2022flamingo} expands large language models to process texts, images, and videos for various visual-linguistic tasks like image captioning and OCR.
In computer vision, VisualPrompting~\cite{bar2022visual} specializes in infographics and images, while Painter~\cite{wang2023images} employs inpainting techniques on masked images for in-context training to solve seven diverse tasks including depth estimation, keypoint estimation, image enhancement, and so on. SegGPT~\cite{wang2023seggpt} explores in-context segmentation tasks in images and videos. Yang et al.~\cite{yang2023exploring} investigate strategies for vision-language model in-context learning on image captioning.
Existing In-Context Learning focuses more on the context of the prompt, using Large Language Models to generate more accurate descriptions, which are then provided to the diffusion model as in-context examples to generate better results. Our method is different. We introduce vision context by using vision latents as the in-context examples.

\section{Method}
\label{sec:method}
In this section, we introduce the problem setup followed by our framework for lifelong few-shot diffusion. We begin by explaining data-free knowledge distillation in Section~\ref{sec:dfkd}, then delve into in-context generation in Section~\ref{sec:ICGen}. The complete pipeline for our lifelong few-shot diffusion is outlined in Figure~\ref{fig:whole}.

\subsection{Problem Set-up}
Lifelong few-shot generation seeks to create a machine learning algorithm capable of continually learning new concepts from a small number of training examples, all while preserving knowledge of previously acquired concepts. This approach typically involves a series of sequential learning sessions. As the model progresses through subsequent sessions, the training datasets from earlier sessions become unavailable. However, evaluating the lifelong few-shot algorithm at each session necessitates generating concepts from both the current session and all prior sessions.
Specifically, let ${\mathcal{D}_{\text{train}}^1, \mathcal{D}_{\text{train}}^2, \ldots, \mathcal{D}_{\text{train}}^n}$ denote the training datasets for various learning sessions, with the text prompt for each dataset $\mathcal{D}_{\text{train}}^i$ indicated as $\mathcal{P}^i$. During the $i$th session, the training is exclusively conducted using $\mathcal{D}_{\text{train}}^i$. The evaluation at session $i$ includes the test prompt $\mathcal{P}_{\text{test}}^n$, which integrates text prompts from both the current and all previous sessions, represented as $\mathcal{P}^0 \cup \mathcal{P}^1 \cup \ldots \cup \mathcal{P}^n$. The training dataset $\mathcal{D}_{\text{train}}^i$ for each session typically contains a limited number of data points, often referred to as a $K$-shot training set, $K$ represents the number of training images and is usually no more than 10. Lifelong few-shot generation presents a significant challenge, marked by severe data imbalance and scarcity, which intensifies the issue of knowledge forgetting in lifelong learning.

In this research, we explore the application of text-to-image diffusion models for lifelong few-shot generation tasks. Stable Diffusion, as described in ~\cite{rombach2022high}, conducts the diffusion process within the latent space which encodes the images using autoencoder, denoted as \(E\). The encoder \(E\) converts an input image \(x\) into a spatial latent representation \(z = E(x)\), which a decoder \(D\)  then uses to reconstruct an approximation of the input image \(D(E(x)) \approx x\). The conditioning vector, corresponding to a specific conditioning prompt \(p\), is represented as \(c(p)\). 
The denoising diffusion network is trained to minimize this loss function: 
\begin{equation}
\label{eq:dmloss}
L_{DM} = \mathbb{E}_{z \sim E(x), p, \epsilon \sim \mathcal{N}(0,1), t} ||\epsilon - \Phi_\epsilon (z_t, t, c(p))||_2^2,
\end{equation}
where \(\Phi_\epsilon\) is the denoising network to estimite the noise \(\epsilon\) to be removed in the noised latent \(z_t\) at timestep \(t\) with condition \(c(p)\). 
During inference, a latent variable \(z_T\) is initially sampled from random noise \(\mathcal{N}(0,1)\). This is iteratively denoised using \(\Phi_\epsilon\) to yield a latent \(z_0\), which the decoder \(D\) then transforms into the final generated image \(x' = D(z_0)\).

\begin{figure}[t]
    \centering
    \includegraphics[width=1.0\linewidth]{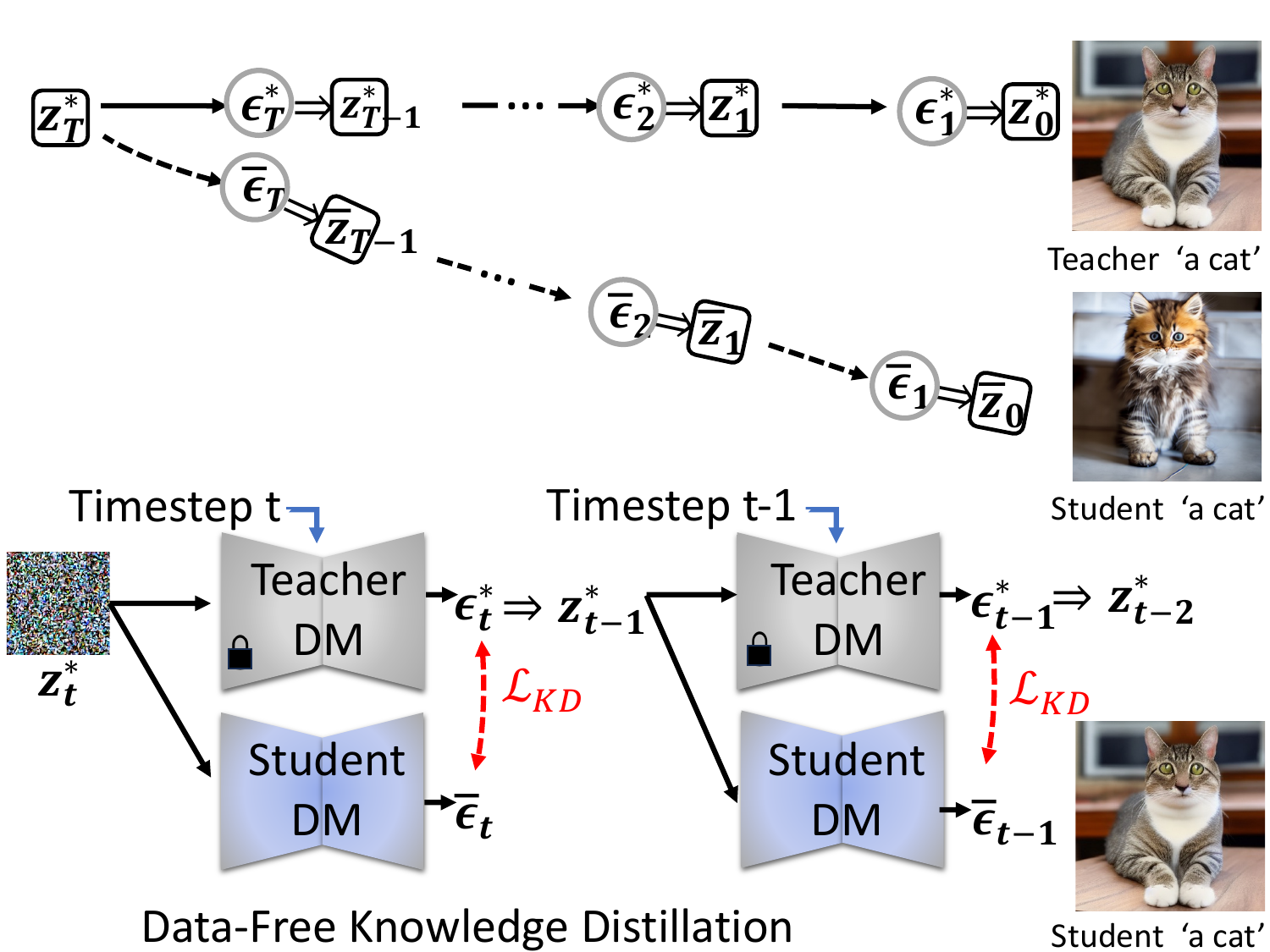}
    \caption{Data-free knowledge distillation. Top: the reverse diffusion process for the behavior of the teacher and student models without distillation. The student model exhibits a tendency to forget the relevant ``cat'' concept and can only generate ``$V_1$ cat'' instead. Bottom: Data-free knowledge distillation for timestep $t$ and timestep $t-1$. For each timestep $t$, we initiate the diffusion process from the identical last teacher latent ${z}_{t}^*$ to prevent discrepancies from accumulating in the teacher and student model outputs. This process involves applying the knowledge distillation loss between the teacher model's noise prediction and the student model, serving to regulate the student model while preserving the relevant concept. }
    \label{fig:kd}
\end{figure}

\subsection{Data-Free Knowledge Distillation for Relevant Concepts Forgetting}
\label{sec:dfkd}

 Fine-tuning the stable diffusion model using new session few-shot training data might result in forgetting the relevant concept in the pre-trained model and the previously learned knowledge from earlier sessions. Knowledge distillation~\cite{hinton2015distilling, Ze_2023} serves as an efficient method to capture the previous model distribution. Given that the original concept is trained on a substantial volume of data that is challenging to access and expensive to incorporate into joint training with few-shot data, we propose Data-Free Knowledge Distillation (DFKD) to acquire the original knowledge during new session model training.

Knowledge Distillation (KD) involves transferring knowledge from a teacher network \(T(x,\theta)\) to a student network \(S(x,\theta_s)\), where \(x\) denotes the input and \(\theta\) signifies the network parameters. The main objective is for the student \(S\) to emulate the behavior of the teacher \(T\) by matching the distribution that the teacher produces on the training data \(x \sim \mathcal{D}_{\text{teacher-train}}\). In our data-free knowledge distillation approach, we eliminate the necessity for \(\mathcal{D}_{\text{teacher-train}}\). Utilizing stable diffusion's ability to generate images from random noise, we implement \(T_\tau\) distillation steps, allowing our teacher model to transfer knowledge to the student model using random noise as input.

The teacher model is defined as \(\Phi_{\epsilon}^*(z_{\tau}^*, \tau, c(p^r))\), where \(z_{T_\tau}^* \sim \mathcal{N}(0,1)\), \(T_\tau\) denotes the distillation steps, and \(p^r \sim P^r\) is the sampled regularization prompts. The student model \(\Phi_\epsilon^{S}(z_{\tau}^*, \tau, c(p^r))\) takes the same random noise as input. The student's training objective is to minimize the distillation loss:
\begin{equation}
\label{eq:kdloss}
L_{KD} = \mathbb{E}_{z_{\tau}^*, p^r, t} ||\Phi_\epsilon^* (z_{\tau}^*, \tau, c(p^r)) - \Phi_\epsilon^{S} (z_{\tau}^*, \tau, c(p^r))||_2^2.
\end{equation}

During the diffusion model training stage, data-free knowledge distillation takes place simultaneously. By incorporating an additional denoising process during training, the diffusion model enhances the knowledge distillation process without the need for real data. The original diffusion model training loss and the newly introduced knowledge distillation loss are applied together, allowing the model to learn new concepts while maintaining knowledge from previous learning stages.
\begin{equation}
\label{eq:trainloss}
L_{train} = L_{DM} + \lambda L_{KD}.
\end{equation}
where \(\lambda\) indicate the strength of the knowledge distillation loss.
Algorithm~\ref{alg:lfs-diffusion} shows the detailed process to train the diffusion model and apply data-free knowledge distillation during training.

As depicted in Figure~\ref{fig:kd}, 
the reverse diffusion process starts with the identical random noise ${z_T^*}$ for both the teacher and student models.
If we consistently denoise along the student diffusion trajectory, a well-learned student model will tend to forget the original concept of ``cat'', resulting in producing all cats look like ``${{V}_1}$ cat''. 
After introducing our DFKD loss $L_{KD}$, The student model is capable of learning the relevant ``cat'' concept.

\begin{algorithm}[t]
\caption{Diffusion model training and Data-free knowledge distillation.
}
\begin{algorithmic}[1]
\REQUIRE Few shot training data $\mathcal{D}_{\text{train}}^i$, Distillation prompt $\mathcal{P}^r$, last session model ${\Phi_\epsilon^{S^{i-1}}}$, distillation scheduler $\sigma_\tau$ with sampling steps $T_\tau$
\label{alg:lfs-diffusion}
\ENSURE A  current session trained diffusion model ${\Phi_\epsilon^{S^i}}$.
\WHILE{not done} 

\STATE $z_0  \sim  \mathcal{D}_{train}^i$, $p \sim \mathcal{P}^i$, $p^r \sim \mathcal{P}^r   \leftarrow$ Encode training image and prompt, Sample regularization prompt  
\STATE ${\Phi_\epsilon^{S^i}} = \Phi_\epsilon^{S^{i-1}}  \leftarrow$ Init current session model
\STATE $\epsilon \sim \mathcal{N}(0,1)      \leftarrow$ Sample random noise
\STATE $z_t \leftarrow \sigma(z_0, t, \epsilon) \leftarrow $ Add noise to training data
\STATE $\epsilon_t = {\Phi_\epsilon^{S^{i}}(z_t, t, c(p))} \leftarrow$ Predict noise added
\STATE $L_{DM} = \mathbb{E}_{z, p, \epsilon, t} ||\epsilon - \epsilon_t||_2^2$
\STATE $z_{T_\tau}^* = \epsilon \leftarrow$ Set teacher model input as random noise
    \FOR{distillation timestep $\tau$ \textbf{in} reversed(range$(T_\tau)$)}
        \STATE $\epsilon_\tau^* = {\Phi_\epsilon^{S^{i-1}}(z_{\tau}^*, \tau, c(p^r))} \leftarrow$ Predict noise by teacher
        \STATE $\epsilon_\tau = {\Phi_\epsilon^{S^{i}}(z_{\tau}^*, \tau, c(p^r))} \leftarrow$ Predict noise by student
        \STATE $L_{KD} = \mathbb{E}_{z_\tau, p^r \sim \mathcal{P}^r, \epsilon_\tau, \tau} ||\epsilon_\tau^* - \epsilon_\tau||_2^2$
        \STATE $\mathcal{L}_{train} = L_{DM} + \lambda  L_{KD} \leftarrow$ Eq(~\ref{eq:trainloss})
        \STATE Optimize ${\Phi_\epsilon^{S^i}}$ with $\mathcal{L}$
        \STATE $z^*_{\tau-1} \sim \sigma_\tau(z_{\tau}^*, \tau, \epsilon_\tau^*) \leftarrow$ Get previous timestep teacher latent for next loop
    \ENDFOR
\STATE $ V_i \leftarrow z_0 $ Sample current session vision context
\ENDWHILE
\end{algorithmic}
\end{algorithm}

\begin{figure}[t]
    \centering
    \includegraphics[width=1.0\linewidth]{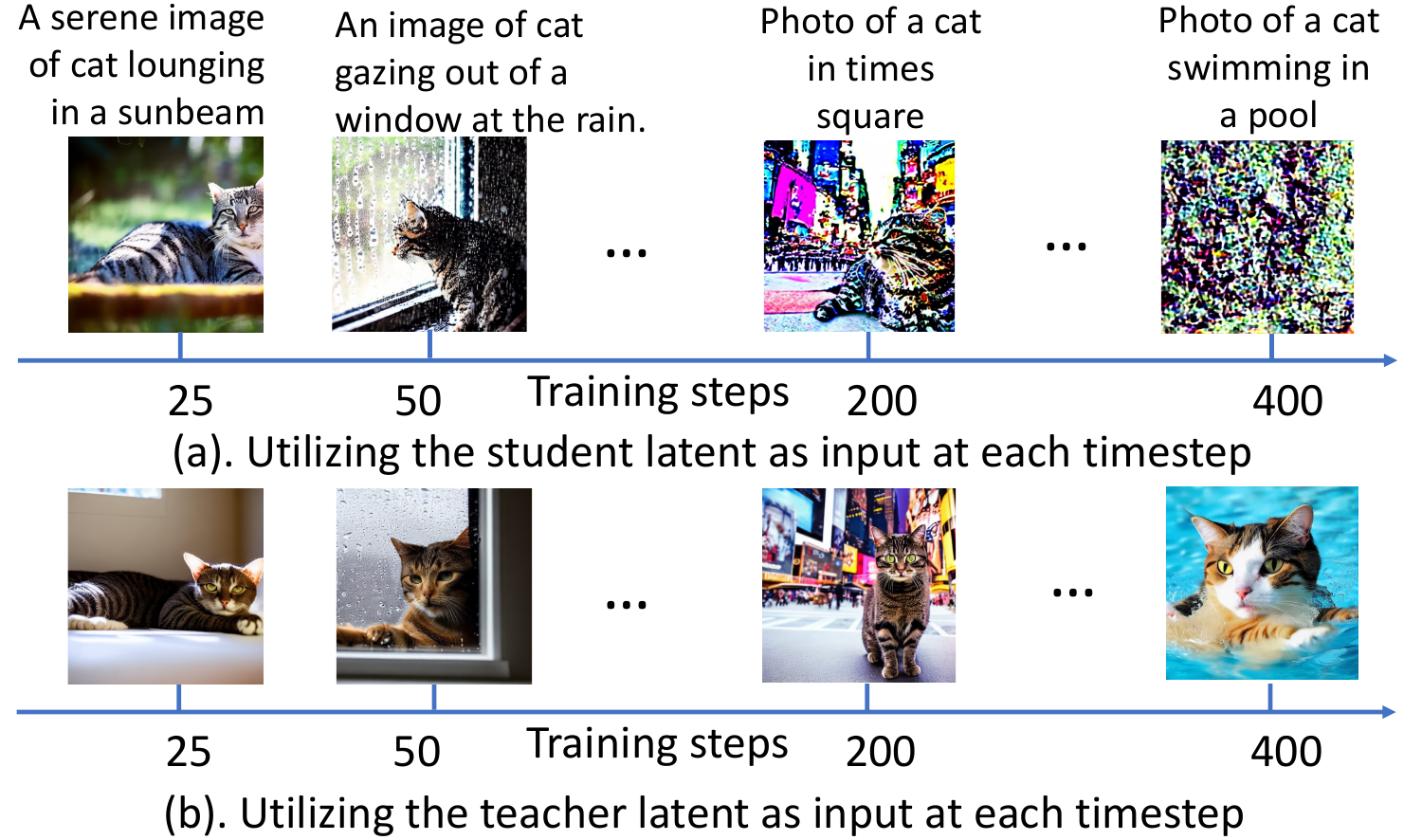}
    \caption{\textbf{Analysis of data-free knowledge distillation along different trajectories.} We present the last DDPM timestep decoding results for the student model. We employ a DDPM scheduler with 25 timesteps during training for distillation. (a)Following the student model reverse diffusion trajectory will cause artifacts when the training steps increase. (b)Utilizing the teacher latent as input for both the teacher and student models at each timestep, followed by distillation between the model outputs, can help eliminate the artifacts problem.   }
    \label{fig:kdabla}
\end{figure}

When we apply knowledge distillation between the sampled teacher instance latent ${z_t^*}$ 
and ${\overline{z}_t}$, 
the model tends to collapse, as shown in Figure~\ref{fig:kdabla}. 
In the early training steps, as ${\Phi}_{\epsilon}$ learns a little information about the new concept, 
it can still generate the original concept effectively. When the training steps increased, distilling the instance $z$ caused these artifacts in the reverse diffusion process. 
This could be due to the error accumulation in the early steps. Therefore, we choose to distill the noise prediction instead of the noisy latent. 
At each timestep $t$, the student model uses the same noisy latent input as the teacher model to prevent the accumulation of artifacts.
Our proposed data-free knowledge distillation between the model noise predictions can prevent the forgetting of the original concepts.

\begin{algorithm}[t]
\caption{Vision in-context inference.
}
\begin{algorithmic}[1]
\REQUIRE $\mathcal{P}_{test}^i$ for session $S^i$, vision context $V_i$, inference scheduler with inference timestep T, the strength of noise added to the vision content $s$.
\label{alg:lfs-icgen}
\ENSURE Generated images for all the seen tasks $X'=\{x'_1,x'_2,\dots,x'_i\}$
    \FOR{session $i$ \textbf{in} $(1,...,S^i)$}
        \STATE $\epsilon \sim \mathcal{N}(0,1) $
        \STATE $ v_i \sim (V_i|p_{test}^i) \leftarrow$ Sample vision context given test prompt
        \STATE $T' = T \times s$
        \STATE $z'_t \leftarrow \sigma(v_i, T', \epsilon) \leftarrow $ Set inference model input as $v_i$ adding random noise at timestep $T'$
        \FOR{timestep $t$ \textbf{in} reversed(range$(T')$)}
            
            \STATE $\epsilon_i = {\Phi_\epsilon^{S^{i}}(z'_t , t, c(p_{test}^i))}$
            \STATE $z'_{t-1} \sim \sigma_I (z'_t, t, \epsilon_i )$
        \ENDFOR
        \STATE $x'_i = \mathcal{D}(z'_0)$
    \ENDFOR
\end{algorithmic}
\end{algorithm}

\subsection{In-Context Generation for Previous Concepts Forgetting}
\label{sec:ICGen}
Our Vision In-Context Generation (ICGen) methodology is structured into two pivotal stages: vision context selection and vision context-guided inference. The process of vision context selection occurs strategically at the culmination of training for each new session, involving the storage of a randomly chosen vision latent within the vision context bank $V_i=\{z_0^{S^1}, z_0^{S^2},...z_0^{S^i}\}$ (as visually represented in Figure~\ref{fig:whole}). Subsequently, during the vision context-guided inference phase, the system strategically selects the corresponding session's vision context based on the provided test prompt $P_{test}^i$. This chosen vision context, combined with random noise, guides the image generation process, as detailed in Algorithm~\ref{alg:lfs-icgen}. In the inference stage, the model utilizes the input of the noisy vision context and the test prompt $p_{test}^i$ to steer the denoising process, optimizing the generation of contextually consistent images. The degree of impact exerted by the vision context is modulated by a specific value $s$, wherein this parameter's range between 0.0 and 1.0 (i.e., $s \in [0,1]$) regulates the quantity of noise incorporated into the visual context. When the value trends toward 1.0, it allows for greater variability but might also result in images potentially lacking semantic coherence within the given visual context. For our Vision In-Context Generation framework, we opt for an intuitive value of $s=0.8$ to balance flexibility and contextual coherence in image generation processes.

\section{Experiment}
\label{sec:experiment}

In this section, we evaluate the performance of our proposed LFS-Diffusion model using the well-known text-to-image customization benchmark datasets, including CustomConcept101~\cite{kumari2022customdiffusion} and the DreamBooth~\cite{ruiz2023dreambooth} datasets, which are well-suited for evaluating few-shot learning and personalized image generation tasks. We begin by detailing the experimental setup and dataset statistics. This is followed by comprehensive experiments to validate the effectiveness of each component in our design and analyze their characteristics. Finally, we compare our framework against state-of-the-art methods on the benchmark dataset to highlight its relative performance.

\subsection{Datasets}
\textbf{CustomConcept101.} CustomConcept101~\cite{kumari2022customdiffusion} is a dataset designed for customized diffusion generation, featuring 101 concepts with 3-15 images per concept and 20 evaluation prompts for each. We adapted the CustomConcept101 dataset to fit a lifelong few-shot task framework, creating five-session tasks to evaluate the performance of lifelong few-shot generation models. The evaluation prompts for each session will encompass all 20 new concept prompts and the prompts from previous sessions. Thus, in session 5, there will be a total of 100 evaluation prompts. This paper introduces two sequences of tasks. The first sequence comprises personal items like a cat, chair, table, flower, and wooden pot. The second sequence includes a personal bike, a specific lighthouse scene, a personal barn, a specific waterfall, and a personal garden view. Our ablation study is conducted on the first sequence.

\textbf{DreamBooth.}
DreamBooth Dataset is a personalized image generation dataset introduced by Ruiz et al.~\cite{ruiz2023dreambooth} The dataset consists of 30 subjects across 15 distinct categories. Each subject has between 4 to 6 images, taken under varying conditions, in different environments. The experiments are designed to simulate a lifelong learning scenario, where the model encounters new tasks in sequential sessions. 
\begin{figure}[tp]
    \centering
    \includegraphics[width=1.0\linewidth]{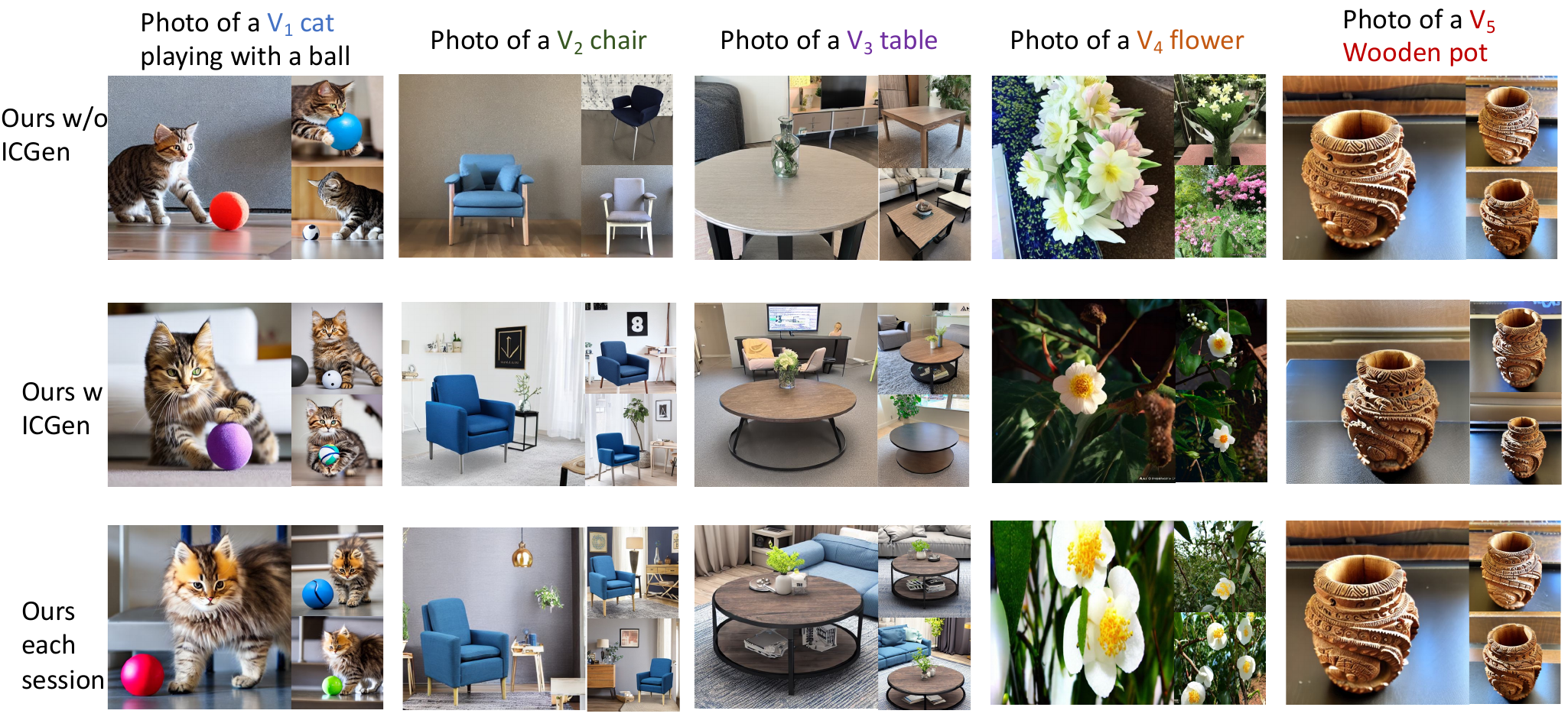}
    \caption{\textbf{Analysis of in-context generation} We present the last session generation results in the first 2 rows. The last row is the generation for each prompt after each session. As we can see, in-context generation helps preserve more old knowledge in lifelong few-shot generation tasks.   }
    \label{fig:abicgen}
\end{figure}

\subsection{Implementation details}
\label{subsection}

We fine-tune the attention layer key and value weights in the SD U-Net, freezing the remaining model components to prevent overfitting in new session training. Initially, the pre-trained stable diffusion model acts as the teacher for the first few-shot learning session. For subsequent sessions, the diffusion model trained in the previous session is used as the teacher model.
ChatGPT~\cite{gpt3} generates our distillation prompts, some adapted from CustomConcept101 datasets, totaling 20 prompts per session for data-free distillation. Our diffusion model is trained for 1000 steps with a learning rate of \(1 \times 10^{-5}\) and a batch size of 1. We incorporate a DDPM scheduler with 25 timesteps for training distillation. In-context generation involves a strength of $s=0.8$ and a 50-step DDPM sampler with a guidance scale of $6$. We run our experiment on two RTX A6000 GPUs.~\cite{kumari2022customdiffusion} takes 110 seconds to run through all training data (including the 200 regularization images), while our model runs through all the training data in just 18 seconds including the data-free distillation process. For new concepts training over 1000 steps, ~\cite{kumari2022customdiffusion} requires 11 minutes, while ours completes in just 9 minutes.

\indent\textbf{Evaluation metrics:} To ensure a fair evaluation of generation performance within our lifelong few-shot learning framework, we generate 50 images for each prompt, yielding a total of 1000 images per concept. Consequently, by session 5, we amass 5000 images for evaluation.
\begin{table}[ht]
  \caption{Ablation on data-free knowledge distillation. In comparing the CLIP Text-Alignment (TA) scores for relevant concepts generated in the final session, we used 100 prompts for the 5 relevant concepts to generate 500 images for each method. Our approach, employing data-free knowledge distillation, exhibits the highest TA among all methods. This result signifies that our method effectively mitigates Relevant Concepts Forgetting (RCF) in contrast to the baseline methods.
  }
  \label{table:comparison}
  \centering
  \begin{tabular}{@{}l|c@{}}
    \hline
   Model & Text-Alignment$\uparrow$ \\
    \hline
	DreamBooth-FT~\cite{ruiz2023dreambooth} & $0.771$    \\
	Custom-FT~\cite{kumari2022customdiffusion} & $0.799$    \\
	Custom-LwF~\cite{kumari2022customdiffusion,li2017learning} & $0.802$    \\
    \hline
        Ours without DFKD & $0.800$   \\	
        Ours with DFKD & $\textbf{0.811}$   \\
  \hline
  \end{tabular}
\end{table}
Following Custom Diffusion~\cite{kumari2022customdiffusion} we include CLIP~\cite{radford2021learning} Image-Alignment(IA) to gauge generation quality, using the target images from the training data in each session. 
For assessing Relevant Concepts Forgetting (RCF), we generate relevant concepts instead of new ones and compute the Text-Alignment (TA) in the last session model. Text Alignment (TA) measure the accuracy of the generated images in relation to the prompts. Evaluating TA for images generated on new concepts might not be ideal for assessing image quality because the CLIP model lacks training on these new concepts and struggles to recognize the newly introduced objects accurately.  To appraise Previous Concepts Forgetting (PCF), we introduce an \textbf{Image-Alignment Dropping} (IAD) metric: 
\begin{equation}
\text{IAD} = \frac{1}{n-1}\sum_{i=1}^{i=n-1}(\text{IA}_i-\text{IA}_n)/(\text{IA}_i)\times100, 
\end{equation}
where $\text{IA}_i$ represents the Image alignment for new concepts learned in session $i$ and $\text{IA}_n$ denotes the IA for the previously learned concept in the last session $n$. Image Alignment Dropping (IAD) is used to evaluate the degree of knowledge retention, specifically how well the model retains concepts across different sessions.
Additionally, a user study was conducted to gather subjective feedback on the quality and consistency of the generated images, providing a practical assessment of the model's performance in real-world applications.

The optimization-based multi-concept generation method outlined in~\cite{kumari2022customdiffusion} (Custom-Optimize) preserves model weights for each new concept and optimizes these distinct weights to generate images for both old and new concepts. However, this approach contradicts the setting of lifelong learning, as it requires saving the old model for each previous concept. Therefore, we will not discuss or compare the Custom-Optimize method in this work. Moreover, our model is also capable of multi-object generation given user-defined layout.
\begin{table*}[ht]
\caption{Image-Alignment results for each session to generate new concepts' images for the first sequence including a cat, chair, table, flower, and wooden pot. The training images are used as the target image for each concept.}
\label{table:ta}
\centering
\begin{tabular}{c|ccccc}
\hline
\multirow{2}{*}{Method} & \multicolumn{5}{c}{Image-Alignment $\uparrow$ } \\
                        & Session 1& Session 2 & Session 3 & Session 4 & Session 5 \\ \hline
DreamBooth-FT ~\cite{ruiz2023dreambooth} & 0.833 & 0.861& \textbf{0.895} & 0.709& \textbf{0.835} \\
Custom-FT~\cite{kumari2022customdiffusion}    & 0.829 & 0.817& 0.806 & 0.687 & 0.809 \\
Custom-LwF~\cite{kumari2022customdiffusion,li2017learning}    & 0.830 & 0.802 & 0.786 & 0.688 & 0.800  \\ \hline
Ours           & \textbf{0.859} & \textbf{0.863} & 0.806 & \textbf{0.752 }& 0.763  \\ \hline
\end{tabular}

\end{table*}
\begin{table*}[ht]
\caption{Comparison of memory storage costs for different methods at the first and last sessions. Our method achieves a 12.6\% memory saving compared to the Custom-FT and Custom-LwF methods.}
\label{table:mem_an}
\centering
\begin{tabular}{c|llllllll}
\hline
\multirow{3}{*}{Method} &
  \multicolumn{8}{c}{Memory} \\ \cline{2-9} 
 &
  \multicolumn{4}{c|}{Session 1} &
  \multicolumn{4}{c}{Session 5} \\ \cline{2-9} 
 &
  \multicolumn{1}{l|}{\begin{tabular}[c]{@{}l@{}}relevant \\ concept\end{tabular}} &
  \multicolumn{1}{l|}{\begin{tabular}[c]{@{}l@{}}vision\\ context\end{tabular}} &
  \multicolumn{1}{l|}{model} &
  \multicolumn{1}{l|}{total} &
  \multicolumn{1}{l|}{\begin{tabular}[c]{@{}l@{}}relevant \\ concept\end{tabular}} &
  \multicolumn{1}{l|}{\begin{tabular}[c]{@{}l@{}}vision\\ context\end{tabular}} &
  \multicolumn{1}{l|}{model} &
  total \\ \hline
Dreambooth-FT~\cite{ruiz2023dreambooth}  &  11MB &  - &  \multicolumn{1}{l|}{3.3GB} & \multicolumn{1}{l|}{3390.2MB} &  11MB &  - &  \multicolumn{1}{l|}{3.3GB} &  3390.2MB \\
Custom-FT~\cite{kumari2022customdiffusion} &  11MB &  - &  \multicolumn{1}{l|}{74MB} &  \multicolumn{1}{l|}{85MB} &  11MB &  - &  \multicolumn{1}{l|}{74MB} &
  85MB \\
Custom-LwF~\cite{kumari2022customdiffusion,li2017learning} &  11MB &  - &  \multicolumn{1}{l|}{74MB} &  \multicolumn{1}{l|}{85MB} &  11MB &  - &  \multicolumn{1}{l|}{74MB} &
  85MB \\ \hline
Ours &  \multicolumn{1}{l|}{-} &  \multicolumn{1}{l|}{0.063MB} &  \multicolumn{1}{l|}{74MB} &  \multicolumn{1}{l|}{74.063MB} &   \multicolumn{1}{l|}{-} &   \multicolumn{1}{l|}{0.315MB} &  \multicolumn{1}{l|}{74MB} &  74.315MB \\ \hline
\end{tabular}%
\end{table*}

\begin{table}[ht]
\centering
\caption{Training time analysis for different methods on two RTX A6000 GPUs, comparing the duration required to train a single new concept and to process all training data, including the data-free distillation process where applicable.}
\label{table:time_an}
\begin{tabular}{c|c}
\hline
Method         & Training time \\ \hline
DreamBooth-FT ~\cite{ruiz2023dreambooth}  & 15 mins       \\
Custom-FT~\cite{kumari2022customdiffusion}    & 11 mins       \\
Custom-LwF~\cite{kumari2022customdiffusion,li2017learning}   & 12 mins       \\ \hline
Ours           & 9 mins        \\ \hline
\end{tabular}
\end{table}

\subsection{Ablation}
\subsubsection{Ablation on data-free knowledge distillation}
In the beginning, we performed an ablative study on data-free knowledge distillation. The Text-Alignment (TA) scores for relevant concepts generated using both baseline methods and our approach demonstrate that our Data-Free Knowledge Distillation (DFKD) effectively prevents Relevant Concepts Forgetting (RCF). We set the sequential fine-tuning on DreamBooth (DreamBooth-FT)~\cite{ruiz2023dreambooth} and Custom Diffusion (Custom-FT)~\cite{kumari2022customdiffusion} as our baseline method. Additionally, we employ a lifelong learning method called Learning without Forgetting (LwF)~\cite{li2017learning} on Custom Diffusion (Custom-LwF) for lifelong few-shot generation. As shown in Table~\ref{table:comparison}, our approach exhibits the highest TA in comparison to the other methods.

\subsubsection{Ablation on in-context generation}
We conduct the ablation study on in-context generation in Figure~\ref{fig:abicgen} and Table~\ref{table:icgen}. Without using ICGen, comparing to the generation results for our model at each session, a lot of the details for the target concept are missing. For example, the distinctive blue color of the $V_2$ chair learned in session 2 is forgotten, resulting in the generation of chairs with altered colors or shapes. With our proposed ICGen approach, a significantly higher level of detail related to the target concept is retained and preserved.

\begin{table*}[ht]
\centering
\caption{\textbf{Quantitative Analysis of In-Context Generation}. Image-Alignment results for last session to generate old concepts' images for the first sequence including a cat, chair, table, and flower. The training images are used as the target image for each concept. As we can see, in-context generation helps preserve more old knowledge in lifelong few-shot generation tasks.}
\label{table:icgen}
\begin{tabular}{c|cccc}
\hline
\multirow{2}{*}{Method} & \multicolumn{4}{c}{Image-Alignment $\uparrow$ } \\
                        & cat & chair & table & flower  \\ \hline \hline
Ours w/o ICG & 0.744 & 0.718& 0.679 & 0.635 \\
Ours w ICG   & \textbf{0.818} & \textbf{0.846}& \textbf{0.759} & \textbf{0.717}  \\
\hline
\end{tabular}%
\end{table*}

\subsection{Memory and Training Time Analysis} Integrating continual learning with customization offers a dynamic and adaptive approach that allows models to evolve and refine their performance over time. This strategy empowers the system to stay relevant in dynamic environments, where new concepts frequently emerge, ensuring a more sustainable and future-proof model. Such an approach mirrors the real-world nature of knowledge acquisition and application. However, handling each new concept separately may lead to escalating memory issues as the number of tasks increases.

We present our memory analysis in Table~\ref{table:mem_an}. The results indicate that the Dreambooth method requires the most storage memory. While the Custom-FT and Custom-LwF methods reduce the memory requirements for the model compared to Dreambooth, the memory needed to retrieve relevant concepts remains significant. Our method, leveraging data-free knowledge distillation, significantly conserves memory, outperforming all other approaches.
For approaches based on Custom Optimization, the weights for all concept models must be preserved. The cumulative memory requirements for the models of five seen concepts in session 5 would be $74\text{MB} \times 5 = 370\text{MB}$. With an additional $11\text{MB}$ required for storing relevant concepts in session 5, the total memory storage would amount to $381\text{MB}$, representing a significant increase compared to the other methods.

We conducted our experiments using two RTX A6000 GPUs. As detailed in Table~\ref{table:time_an}, training Dreambooth for a single new concept takes approximately 15 minutes. Custom-FT~\cite{kumari2022customdiffusion} processes all training data, including 200 regularization images, in 110 seconds, whereas our model completes the training through all data in just 18 seconds, including the data-free distillation process. When training over 1000 steps for new concepts, Custom-FT~\cite{kumari2022customdiffusion} requires 11 minutes, while our method completes the task in only 9 minutes. Our approach provides a significant advancement in the field, combining speed, efficiency, and quality in training models for continually learning new concepts.

\begin{table*}[t]
\caption{Image-Alignment Dropping(IAD) results for each session to generate previous concepts' images for the first sequence including cat $\rightarrow$ chair $\rightarrow$ table $\rightarrow$ flower $\rightarrow$wooden pot. The training images are used as the target image for each concept.}
\label{table:sota}
\centering
\begin{tabular}{c|cccc}
\hline
\multirow{2}{*}{Method} & \multicolumn{4}{c}{Image-Alignment Dropping (IAD) \% $\downarrow$ } \\
                        & Session 2 & Session 3 & Session 4 & Session 5 \\ \hline
DreamBooth-FT ~\cite{ruiz2023dreambooth} & 5.01 & 8.91 & 11.50 & 14.54 \\
Custom-FT~\cite{kumari2022customdiffusion}    & 6.94 & 7.68 & 8.92 & 10.20 \\
Custom-LwF~\cite{kumari2022customdiffusion,li2017learning}    & 7.23 & 6.67 & 8.28 & 8.71  \\ \hline
Ours           & \textbf{4.58} & \textbf{2.84} & \textbf{3.40 }& \textbf{4.29 } \\ \hline
\end{tabular}

\vskip -1em
\end{table*}

\subsection{Comparison with the State-of-the-Art Methods}
\label{section:sota}
Lastly, we evaluate our model's performance against state-of-the-art methods, including Dreambooth-FT~\cite{ruiz2023dreambooth}, Custom Diffusion (Custom-FT)~\cite{kumari2022customdiffusion}, and 
Custom Diffusion with Learning without Forgetting ~\cite{li2017learning} (Custom-LwF) on the benchmark dataset, CustomConcept101 and DreamBooth. 
Our results demonstrate that the proposed LFS-Diffusion framework outperforms existing methods, such as DreamBooth-FT and Custom Diffusion, on both TA and IAD metrics. For instance, our method achieves the highest TA scores, particularly in later sessions, where catastrophic forgetting is most problematic for baseline models. Additionally, the IAD metric shows that our method retains significantly more knowledge of previously learned concepts compared to other approaches.

\begin{figure}[t]
	\centering
	\includegraphics[width=1.0\linewidth]{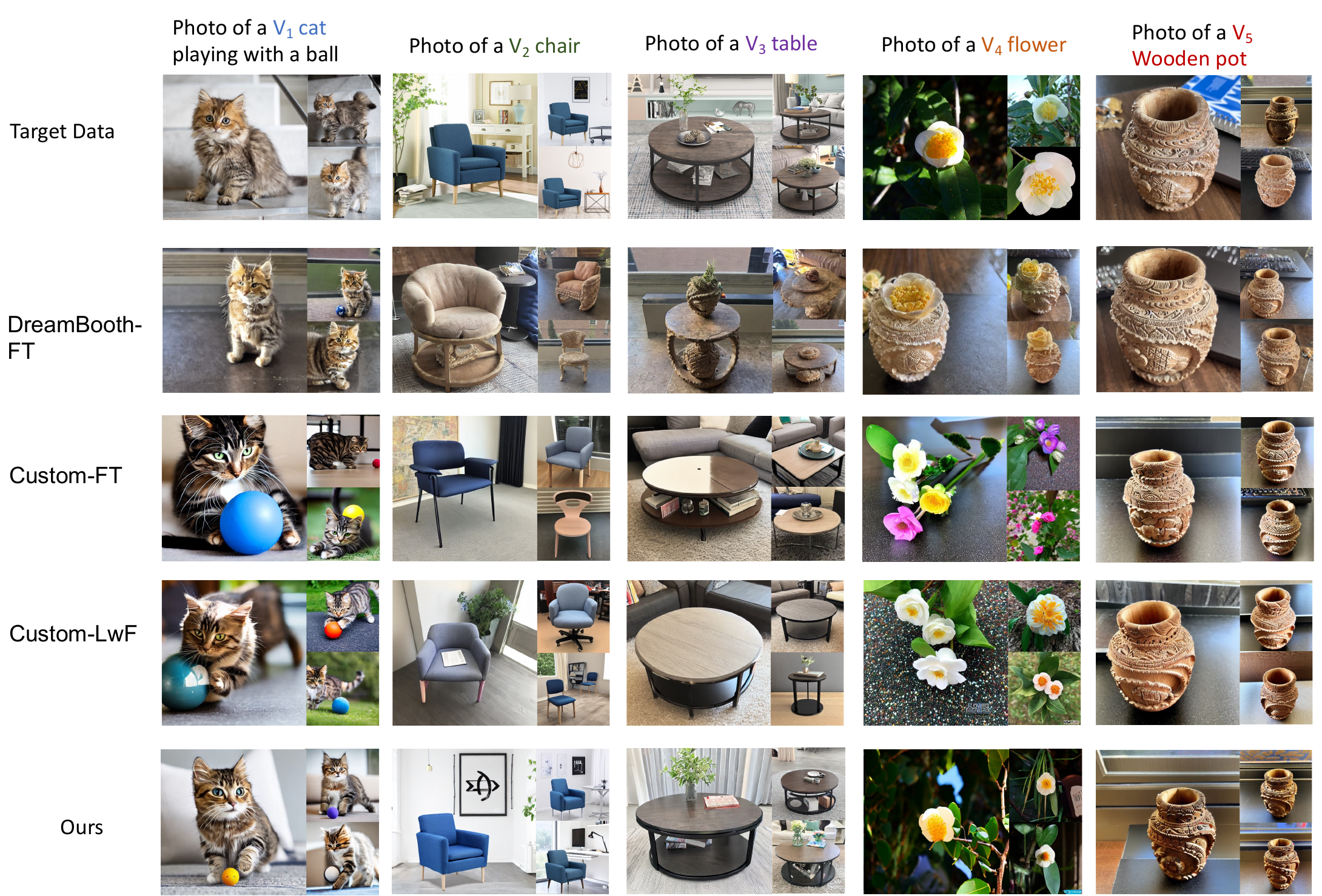}
	\caption{Qualitative results of lifelong few-shot generation at session 5 on CustomConcept101~\cite{kumari2022customdiffusion} dataset selected sequence, $V_1$ cat $\rightarrow$ $V_2$ chair $\rightarrow$ $V_3$ table $\rightarrow$$V_4$ flower $\rightarrow$ $V_5$ wooden pot. The first row illustrates the training data. DreamBooth-FT~\cite{ruiz2023dreambooth} fine-tunes all parameters, resulting in the generation of objects resembling the $V_5$ wooden pot across all concepts. Custom-FT and Custom-LwF ~\cite{kumari2022customdiffusion}  methods exhibit signs of Previous Concepts Forgetting (PCF). For instance, the blue $V_2$ chair learned in the second session is not accurately generated using these methods. our method in the last row showcases high-quality image generation without any signs of forgetting.
	}
	\label{fig:sota}
\end{figure}

\begin{figure}[h]
	\centering
	\includegraphics[width=1\linewidth]{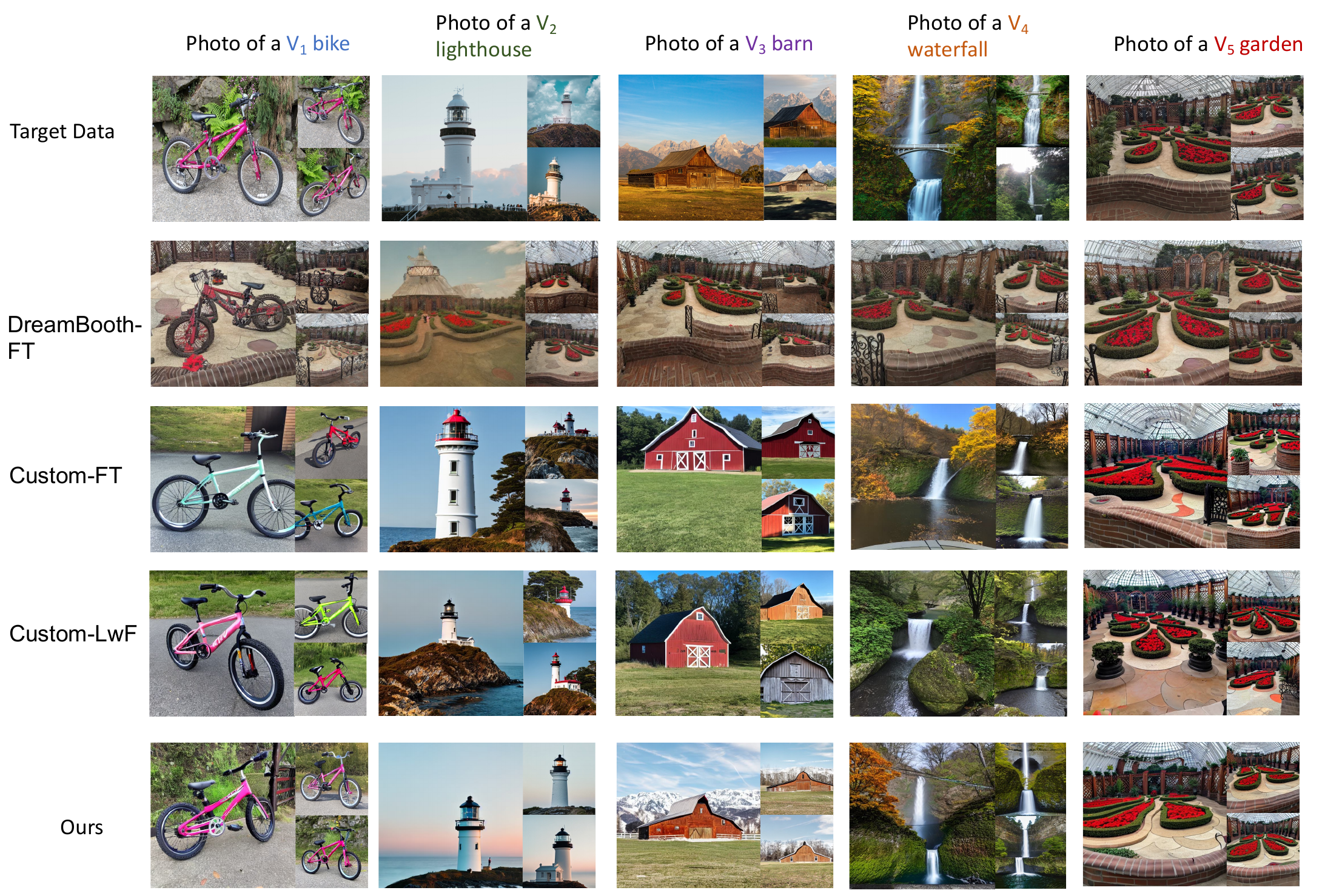}
	\caption{Qualitative results of lifelong few-shot generation at session 5 on CustomConcept101~\cite{kumari2022customdiffusion} dataset selected sequence, $V_1$ bike $\rightarrow$ $V_2$ lighthouse $\rightarrow$ $V_3$ barn $\rightarrow$$V_4$ waterfall $\rightarrow$ $V_5$ garden. The first row illustrates the training data. DreamBooth-FT~\cite{ruiz2023dreambooth} fine-tunes all parameters, resulting in the generation of objects resembling the $V_5$ garden across all concepts. Custom-FT and Custom-LwF ~\cite{kumari2022customdiffusion}  methods exhibit signs of Previous Concepts Forgetting (PCF). For instance, the $V_3$ barn learned in the third session is not accurately generated using these methods. our method in the last row showcases high-quality image generation without forgetting.
	}
	\label{fig:sota_bike}
\end{figure}

We show the results of the first sequence including a cat, chair, table, flower, and wooden pot. we presenting our findings in Figure~\ref{fig:sota} and detailing the Image-Alignment, the Image-Alignment Dropping (IAD) metric in Table~\ref{table:sota}. Our model achieves the lowest image-alignment dropping rate, exceeding state-of-the-art results by 4.42\%.
\begin{table*}[t]
\caption{Image-Alignment Dropping(IAD) results for each session to generate images of previous concepts for sequence $V_1$ bike $\rightarrow$ $V_2$ lighthouse $\rightarrow$ $V_3$ barn $\rightarrow$$V_4$ waterfall $\rightarrow$ $V_5$ garden. The training images are used as the target image for each concept.}
\label{table:sota_bike}
\centering
\begin{tabular}{c|cccc}
\hline
\multirow{2}{*}{Method} & \multicolumn{4}{c}{Image-Alignment Dropping (IAD) \% $\downarrow$ } \\
                        & Session 2 & Session 3 & Session 4 & Session 5 \\ \hline
DreamBooth-FT ~\cite{ruiz2023dreambooth} & 8.24 & 19.51 & 25.34 & 32.42 \\
Custom-FT~\cite{kumari2022customdiffusion}    & 10.90 & 12.09 & 11.56 & 11.29 \\
Custom-LwF~\cite{kumari2022customdiffusion,li2017learning}    & 9.18 & 11.64 & 11.07 & 11.00  \\ \hline
Ours           & \textbf{7.50} & \textbf{5.87} & \textbf{7.94 }& \textbf{7.31 } \\ \hline
\end{tabular}%
\end{table*}
The results of the second sequence including a personal bike, a specific lighthouse scene, a personal barn, a specific waterfall, and a personal garden view in Figure.~\ref{fig:sota_bike}. Next, we calculate the Image-Alignment Dropping (IAD) metric for each method in Table~\ref{table:sota_bike}, allowing for a comparative analysis of their abilities to preserve previous concepts. Our method demonstrates the lowest IAD across all sessions, indicating its superior ability to preserve previous knowledge.

\begin{figure}[t]
\begin{center}
\includegraphics[width=1\linewidth]{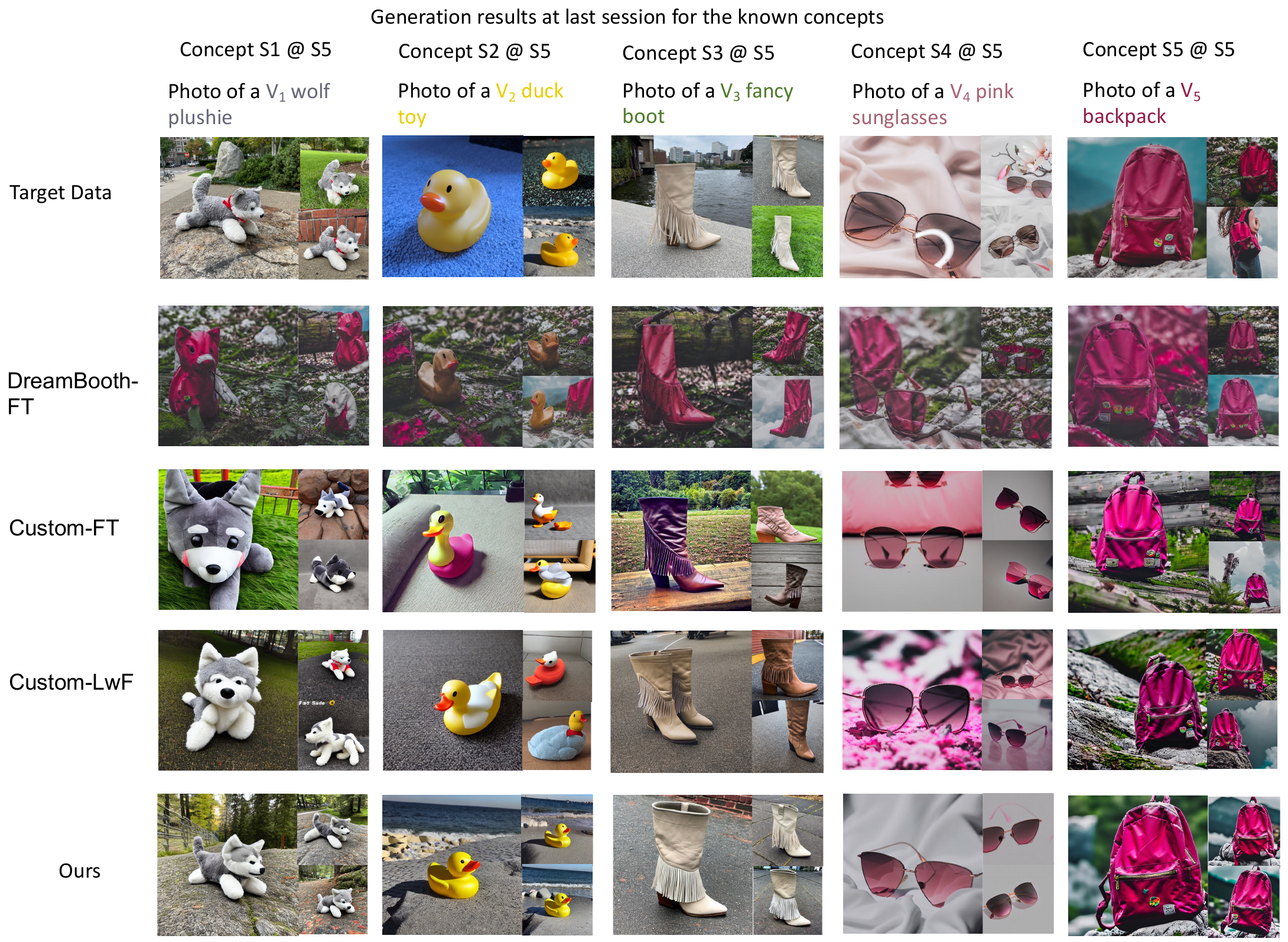}
	\caption{Qualitative results of lifelong few-shot generation at session 5 on DreamBooth~\cite{ruiz2023dreambooth} dataset selected sequence, $V_1$ wolf plushie $\rightarrow$ $V_2$ duck toy $\rightarrow$ $V_3$ fancy boot $\rightarrow$$V_4$ pink sunglasses $\rightarrow$ $V_5$ backpack. The first row illustrates the training data. DreamBooth-FT[~\cite{ruiz2023dreambooth}] fine-tunes all parameters, resulting in the generation of objects resembling the $V_5$ backpack across all concepts. Custom-FT and Custom-LwF ~\cite{kumari2022customdiffusion}  methods exhibit signs of Previous Concepts Forgetting (PCF). For instance, the yellow $V_2$ duck toy learned in the second session is not accurately generated using these methods. our method in the last row showcases high-quality image generation without any signs of forgetting.  }
	\label{fig:sota_new_dataset}
\end{center}
\end{figure}

we add an additional experimental result on the DreamBooth~\cite{ruiz2023dreambooth} dataset for the selected sequence, $V_1$ wolf plushie $\rightarrow$ $V_2$ duck toy $\rightarrow$ $V_3$ fancy boot $\rightarrow$$V_4$ pink sunglasses $\rightarrow$ $V_5$ backpack as shown in Figure.~\ref{fig:sota_new_dataset}. This new dataset,enables a more robust comparison of our method against established techniques. The ability of our model to retain personalized image features across sessions highlights its effectiveness in personalized diffusion tasks, addressing catastrophic forgetting while maintaining fine-grained details.

\begin{figure*}[t]
\begin{center}
\includegraphics[width=1\linewidth]{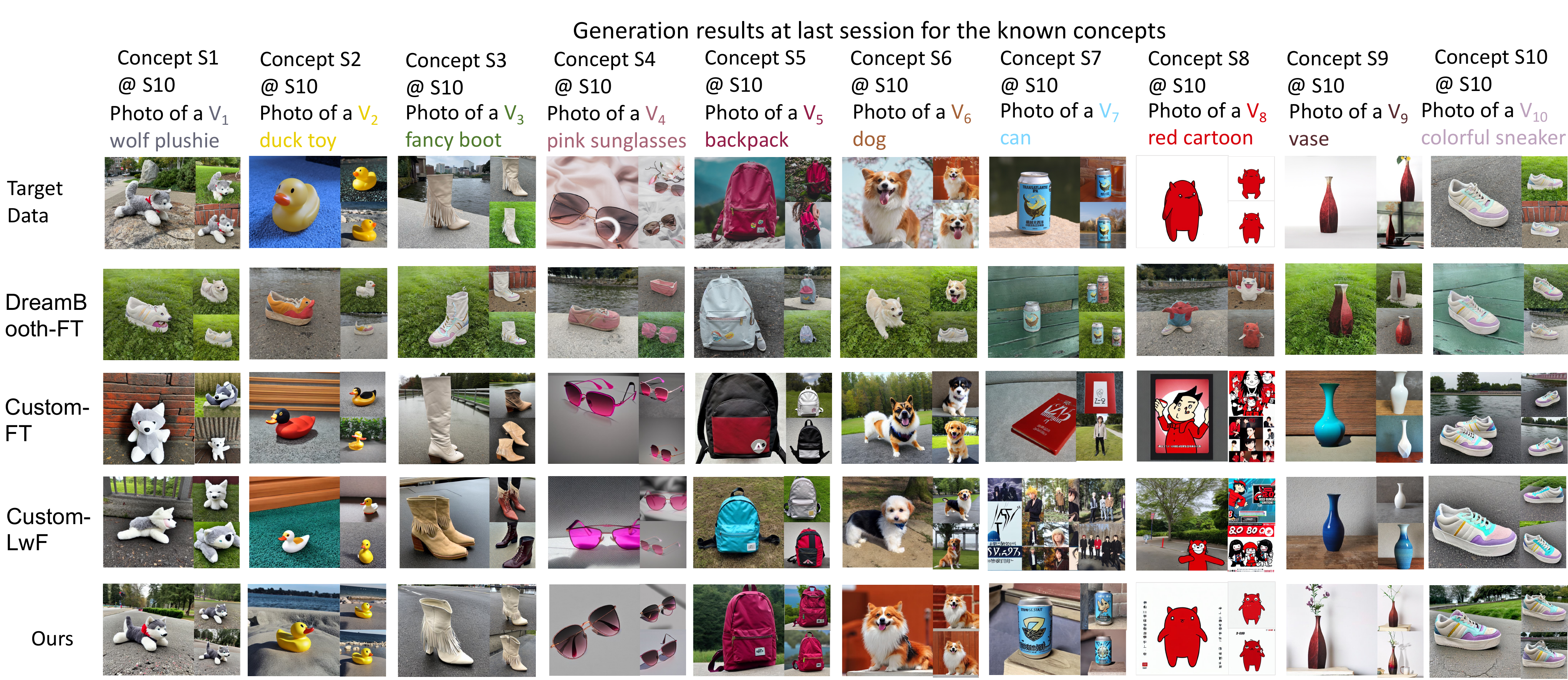}
we also conduct long-term learning experimental result on the DreamBooth [14] dataset for 2 times longer (10 sessions) as shown in Figure.~\ref{fig:sota_new_dataset_long}.

	\caption{Qualitative results of lifelong few-shot generation at session 10 on DreamBooth~\cite{ruiz2023dreambooth} dataset selected sequence, $V_1$ wolf plushie $\rightarrow$ $V_2$ duck toy $\rightarrow$ $V_3$ fancy boot $\rightarrow$$V_4$ pink sunglasses $\rightarrow$ $V_5$ backpack $\rightarrow$ $V_6$ dog $\rightarrow$ $V_7$ can $\rightarrow$$V_8$ red cartoon $\rightarrow$ $V_9$ vase $\rightarrow$ $V_{10}$ colorful sneaker. The first row illustrates the training data. DreamBooth-FT[14] fine-tunes all parameters, resulting in the generation of objects resembling the $V_10$ colorful sneaker across all concepts. Custom-FT and Custom-LwF ~\cite{kumari2022customdiffusion}  methods exhibit signs of Previous Concepts Forgetting (PCF). For instance, the yellow $V_2$ duck toy learned in the second session is not accurately generated using these methods. our method in the last row showcases high-quality image generation without any signs of forgetting.  }
	\label{fig:sota_new_dataset_long}
\end{center}
\end{figure*}

\begin{figure}[h]
	\centering
	\includegraphics[width=1\linewidth]{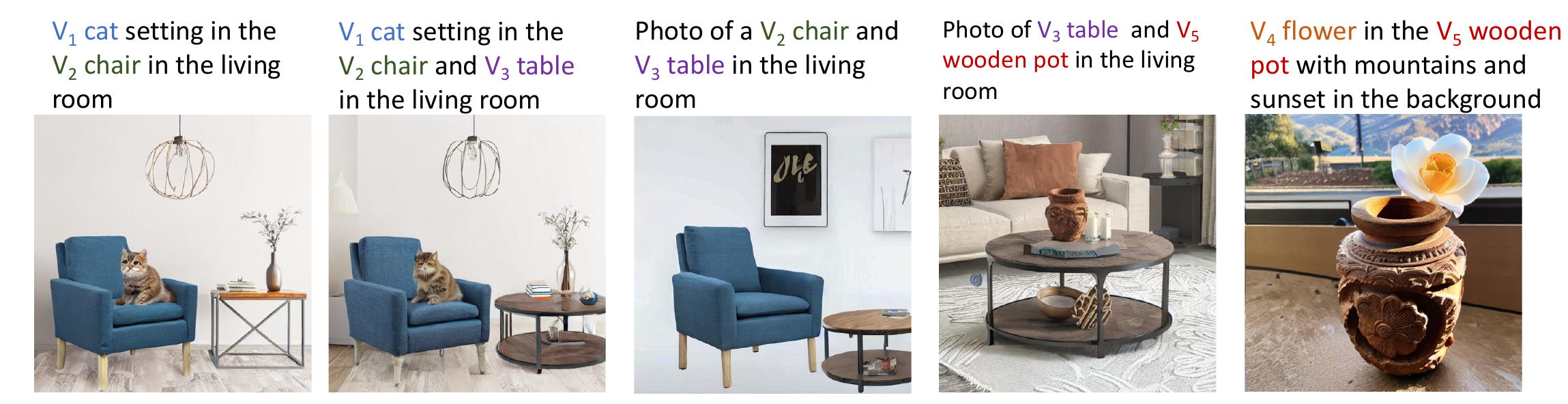}
	\caption{Multi-concept generation results.
	}
	\label{fig:multi-gen}
\vskip -1em
\end{figure}

\subsection{User Study} We also conduct user studies to quantitatively evaluate our method. Participants are given the training image, captions and different models generation results to evaluate which model generates better result from a human perspective. The user study is conducted on 100 images. We run 5 sessions and 10 sessions lifelong learning experiments and generate 100 images for each method. Based on our user studies, it is found that 77.8\% of votes think our method generates better results from a human perspective as shown in Table~\ref{table:user}.

\begin{table*}[ht]
\centering
\caption{\textbf{Quantitative Comparison}. Ours outperform to the other baseline methods in the user study results.}
\label{table:user}
\begin{tabular}{ccccc}
\hline
\multicolumn{1}{c}{ } & DreamBooth\_FT & Custom\_FT & Custom\_LwF & Ours\\ \hline\hline
User study  & 3.34\%     & 5.89\%   & 12.97\%& \textbf{77.80\%}      \\ 
\hline \\
\end{tabular}
\end{table*}

\subsection{Multi-concept Generation}
As mentioned in Section \ref{subsection} of our research, our method extends its capabilities to multi-object generation based on a user-defined layout. This functionality is particularly advantageous for users aiming to craft customized scenes or compositions. By allowing users to specify the desired arrangement of objects, our approach not only facilitates personalized image creation but also enhances the applicability of the tool in design and simulation tasks where specific spatial configurations are critical.

To implement this feature, we adapt the model to interpret and integrate a user-defined layout into its processing workflow. Initially, the model takes a standard single visual context as its input. This context is then dynamically updated to align with the user-provided layout, which acts as a blueprint for the generation process. The layout specifies the precise positions and relational dynamics among the objects, effectively directing the model on where and how to place each element within the scene. This structured approach guarantees that the generated objects meet the user-specified spatial and contextual requirements, thereby improving both the aesthetic and functional aspects of the output.

Leveraging this enhanced visual context, our model performs what we refer to as multi-concept In-Context Generation (ICGen). During this phase, the model utilizes the combined visual context, now enriched with layout information, to generate multiple objects within a single image. This process not only considers the individual characteristics of each object but also their spatial arrangement and mutual interactions as dictated by the layout. By doing so, ICGen produces a cohesive scene where all elements are contextually integrated, reflecting a realistic and harmonious interplay between them, as shown in Figure~\ref{fig:multi-gen}. This method is particularly effective in scenarios requiring complex interactions among multiple elements, such as dynamic scenes in gaming environments or detailed project visualizations in architectural design.

\section{Conclusion}
\label{sec:conclusion}
In this paper, we address the challenges of lifelong few-shot customization in text-to-image diffusion models from two distinct perspectives. Firstly, the implementation of a data-free knowledge distillation method effectively mitigated knowledge forgetting within the model's backbone. Secondly, our paradigm, In-Context Generation (ICGen), empowers the diffusion model by enabling in-context learning during inference, substantially improving few-shot generation and alleviating issues of past concept forgetfulness. Through extensive experiments on the CustomConcepts101 and DreamBooth dataset, our method demonstrated exceptional performance, outperforming baseline models and setting a new benchmark in this field.

{\small
\bibliographystyle{IEEEtran}
\bibliography{references}
}


\newpage

\vfill

\end{document}